\documentclass[10pt,twocolumn,letterpaper]{article}

\usepackage{cvpr}
\usepackage{times}
\usepackage{epsfig}
\usepackage{graphicx}
\usepackage{amsmath}
\usepackage{amssymb}
\usepackage{xcolor}
\newcommand{\myparagraph}[1]{\vspace{0.1em}\noindent\textbf{#1}}
\definecolor{joonscomment}{RGB}{179,93,255}

\definecolor{qianrucom}{RGB}{50,50,255}
\usepackage[breaklinks=true,bookmarks=false]{hyperref}
\usepackage{multirow}
\usepackage{color}
\usepackage{wrapfig}
\usepackage{xspace}
\usepackage{caption}
\usepackage{url}            
\usepackage{booktabs,tabularx}       
\usepackage{tablefootnote}
\cvprfinalcopy 
\def\cvprPaperID{1817} 

\ifcvprfinal\fi
\begin{document}
\pagestyle{plain}
\title{Natural and Effective Obfuscation by Head Inpainting}
\author{Qianru Sun$^{1}\footnotemark[1]$ \quad Liqian Ma$^{2}\thanks{Equal contribution.} $ \quad Seong Joon Oh$^{1}$ \\ Luc Van Gool$^{2,3}$ \quad Bernt Schiele$^{1}$  \quad Mario Fritz$^{1}$ \\
\\
  $^{1}$Max Planck Institute for Informatics, Saarland Informatics Campus \\
  $^{2}$KU-Leuven/PSI, Toyota Motor Europe (TRACE) \quad  $^{3}$ETH Zurich \\
  {\texttt{\{qsun, joon, schiele, mfritz\}@mpi-inf.mpg.de}} \\
  {\texttt{\{liqian.ma,
luc.vangool\}@esat.kuleuven.be}} \\{\texttt{vangool@vision.ee.ethz.ch}}
}

\maketitle

\begin{abstract}
As more and more personal photos are shared online, being able to \emph{obfuscate} identities in such photos is becoming a necessity for privacy protection. People have largely resorted to blacking out or blurring head regions, but they result in poor user experience while being surprisingly ineffective against state of the art person recognizers~\cite{joon16eccv}. In this work, we propose a novel \emph{head inpainting} obfuscation technique. Generating a realistic head inpainting in social media photos is challenging because subjects appear in diverse activities and head orientations. We thus split the task into two sub-tasks: (1) facial landmark generation from image context (e.g. body pose) for seamless hypothesis of sensible head pose, and (2) facial landmark conditioned head inpainting. We verify that our inpainting method generates realistic person images, while achieving superior obfuscation performance against automatic person recognizers.
\end{abstract}

\section{Introduction}

Social media have brought about large-scale sharing of personal photos. While providing great user convenience, such a dissemination can pose privacy threats on users. It is essential to grant users an option to obfuscate themselves out of these photos.
A good obfuscation method for social media photos should satisfy two criteria: \emph{naturalness} and \emph{effectiveness}. For example, putting a large black box over a person may be an effective obfuscation method, but would not be pleasant enough to share with friends. 

\begin{figure}[htp]
  \centering
  \includegraphics[width=0.9\linewidth]{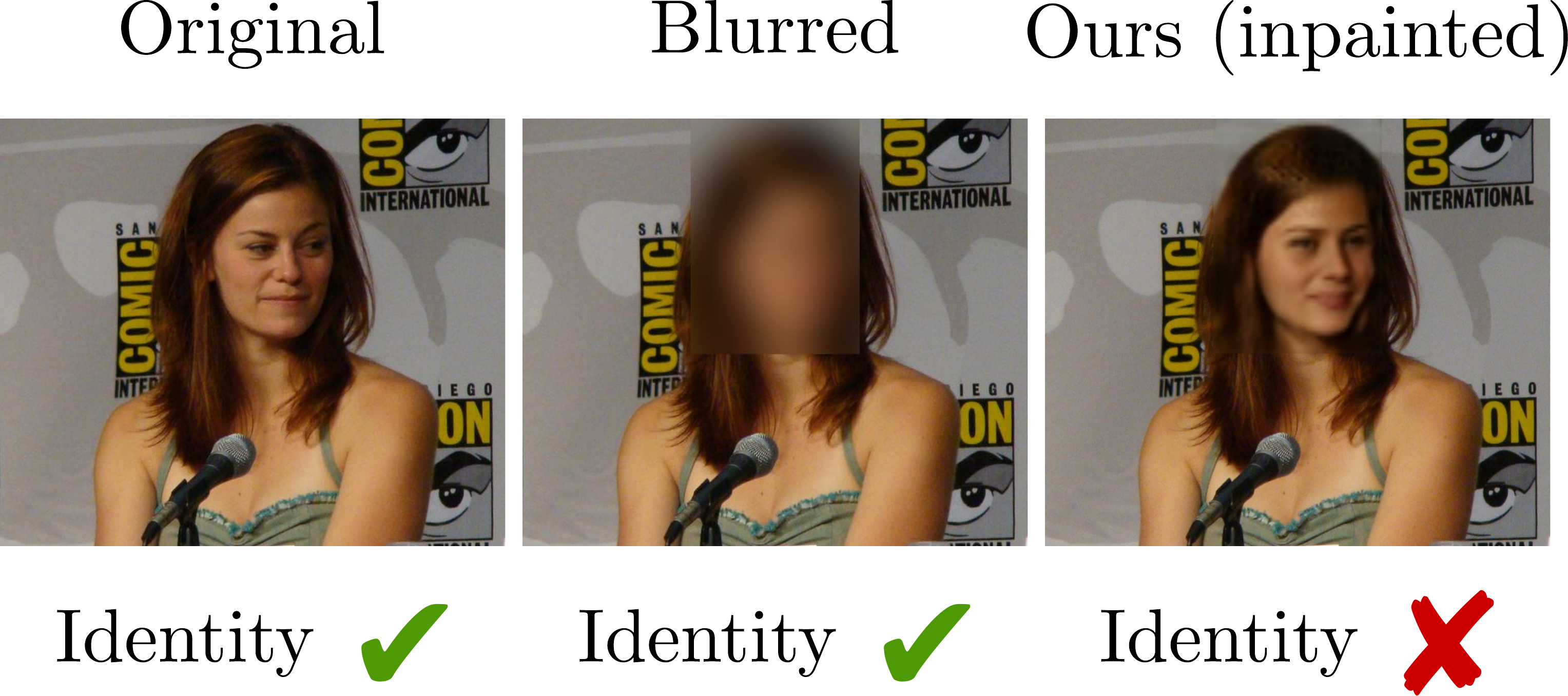}\\
  \caption{Our obfuscation method based on head inpainting generates much more natural patterns than common techniques like blurring, but still results in a more effective identity obfuscation against a recognizer.}
\label{framework_obfusc}
\end{figure}

Previous work on visual content obfuscation can be grouped into two categories: (1) \emph{target-specific} and (2) \emph{target-generic}.  Some papers have proposed \emph{target-specific} obfuscations, ones that are specialized against specific target machine systems, typically relying on adversarial examples~\cite{joon17iccv,Sharif16AdvML}. They yield nearly perfect identity protection with imperceptible changes on the input, but such a performance is guaranteed only against the targetted ones.

On the other hand, \emph{target-generic} obfuscations change the actual appearance of the person such that generic classifier or even humans misjudge the identity. In its most crude form, commonly used obfuscation methods like black eye bar, face blurring, and blacking out head are examples of this type. These common patterns, unfortunately, are neither visually pleasant nor effective against machine systems~\cite{joon16eccv}. This paper proposes a \emph{head inpainting} based approach to the target-generic identity obfuscation problem.

Generating realistic and seamless head inpainting on social media photos is hard. Subjects appear in diverse events and activities, resulting in varied backgrounds and head poses. Meanwhile, current generative face models are limited to frontal \cite{ColeIccv2017} or strictly aligned faces \cite{ZheheLu17}. 

We tackle the problem by factoring it into two stages. First, depending on the input, we detect or generate facial landmarks. In particular, when we have access to the original image, we detect facial landmarks. However, to keep our approach versatile, we also address the more challenging problem of generating facial landmarks from images that have been already obfuscated e.g. by a blacking out the face (called blackhead in the remainder of the paper).  
Then, conditioned on the face landmarks, we inpaint a realistic head that blends naturally into the context. We show that the resulting head-inpainted images mislead machine recognizers. Note that our method supports cases where the original face image is not available; existing head-obfuscated images on the web can be ``upgraded'' to our privacy enhanced head inpainting. 

Key contributions are: 
(1) Novel natural, effective obfuscation methods based on head inpainting;
(2) Novel landmark guided image generation approach for both head visible and blackhead cases in challenging social media photos; 
(3) Novel facial landmark generator that effectively hypothesize realistic facial structures and poses given context in the scenario of blackhead.

This paper is organized as follows. 
In Section~\ref{sec:related_work} we present related works mainly on identity obfuscation and image inpainting.
Section~\ref{sec:method} describes the proposed two-stage framework in detail. 
Section~\ref{sec:experiments} evaluates the presented method in the context of person obfuscation in social media.

\section{Related work}
\label{sec:related_work}

\paragraph{Identity Obfuscation}
A few works from the vision community have analyzed and developed obfuscation patterns for protecting private visual content. 
First, we introduce a line of work on \emph{target-generic} obfuscations that are designed to work against generic automatic person recognizers as well as humans. Oh \etal~\cite{joon16eccv} and McPherson \etal~\cite{mcpherson2016defeating} have analyzed the obfuscation performance of blacking or blurring faces against automatic recognizers. They have concluded that these common obfuscation methods are not only unpleasant but also ineffective, in particular due to the adaptability of convnet-based recognizers \cite{joon16eccv}. More sophisticated approaches have been proposed since then. Hassan \etal~\cite{hassanCVPRW17} have proposed to mask private image content via \emph{cartooning}. Brkic \etal~\cite{brkicCVPRW17} have generated full-person patches to overlay on top of person masks. Similarly, we propose an obfuscation technique based on \emph{head inpainting}. The key difference is that while \cite{brkicCVPRW17} generates persons with uniform poses independent of the context (fashion photos), we naturally blend generated heads with diverse poses into varied background and body poses (social media photos).

For the \emph{target-specific} obfuscations, Oh \etal~\cite{joon17iccv} and Sharif \etal~\cite{Sharif16AdvML} have proposed \emph{adversarial example} based obfuscation techniques. Pros are that the obfuscation patterns are imperceptible for humans and obfuscation performance is superb; cons are that such a performance is only guaranteed for a few targetted machine systems.

\myparagraph{Image inpainting}
In our work, we propose generative adversarial network (GAN) based method to complete head regions based on the context. 
Raymond \etal~\cite{YehCLHD16} and Pathak \etal~\cite{contextEncoder} have also used GANs to generate missing visual contents, conditioning on the context. However, both approaches assume appearance and texture similarity between the missing part and the context. Our approach can generate head inpainting solely from body and scene context, without resorting to any information from the head region. In particular, unlike \cite{YehCLHD16} method which has been applied to aligned face images, our approach can be applied to challenging social media setup in which people appear with diverse poses and backgrounds by taking a two-stage approach.

\myparagraph{Structure guided image generation}
For generating realistic head inpainting that naturally blends into the given body pose and scene context, we have conditioned the inpainting on face landmarks. Some prior work has been devoted to the structure guided image generation; such a guidance has proved very helpful for generating images with complex inner structures (e.g. persons)~\cite{PG2,GP-GAN,WalkerMGH17,S2-GAN,SeGAN,stackGAN,Ma2018}.
Ma \etal~\cite{PG2} embed an arbitrary pose into a reference person image, and then refine the output by decoding more appearance details in the second stage. Alpher \etal~\cite{GP-GAN} use a similar structure embedding method to generate face image with detected facial landmarks on well-aligned face dataset. Walker \etal~\cite{WalkerMGH17} modeled the possible future movements of humans in the pose space, and then used the future poses generated as conditional information to a GAN to predict the future frames of the video.
In~\cite{S2-GAN}, Wang and Gupta propose to first generate a 3D surface normal map from a Gaussian signal and then synthesize images by painting style information on the map. Ehsani \etal~\cite{SeGAN} solve the problem of object occlusion by first predicting the contour of invisible part then generate the appearance inside this contour. The second stage replies on the close visibility same as Context Encoder~\cite{contextEncoder}. 
Cole and Belanger ~\cite{ColeIccv2017} recently introduce an approach face warping manipulation using landmark control on frontal face images. 
Different with these landmark works, our approach can not only generate new landmarks from body context, but also handle the large pose variances in Flickr images.

\section{Head inpainting framework}
\label{sec:method}

We propose a context-driven head inpainting approach. We focus on social media photos which are  challenging due to complex poses and scenarios. To learn an effective head generator from the data, we need strong guidance for which we use facial landmarks. 
Therefore, we factor the head inpainting task into two stages: landmark detection/generation and head inpainting conditioned on body context and landmarks. 

\begin{figure}[htp]
  \centering
  \includegraphics[width=0.99\linewidth]{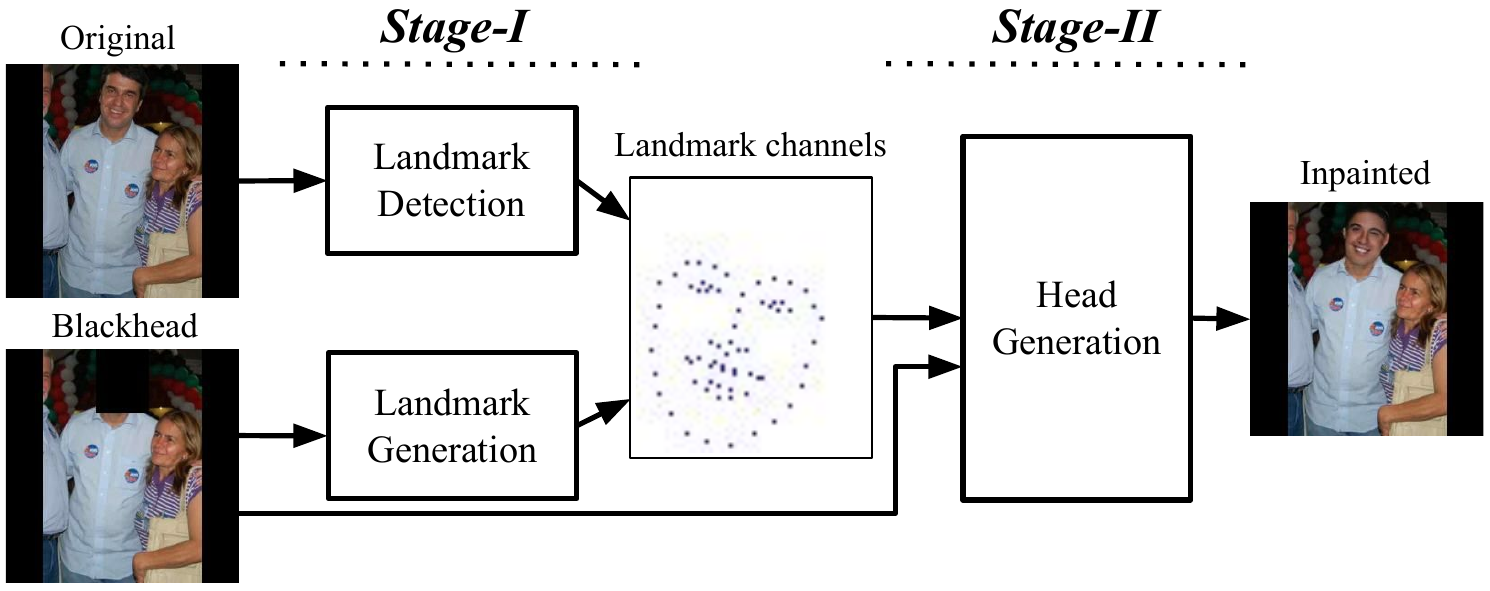}\\
  \caption{Our two-stage head inpainting framework. The input of stage-I is either the original or the blackhead image. The output is the inpainted image.}
\label{framework_main}
\end{figure}

Figure~\ref{framework_main} describes the global view of our two-stage approach. It takes either original or blackhead image\footnote{Blurhead image is another important obfuscation, and it is easily adapted in our approach. We use blackhead image as a default example.} as input, in order to give flexibility to deal with cases where the original images are not available. Given original or head-obfuscated input, stage-I detects or generates landmarks, respectively. Stage-II takes the blackhead image and landmarks as input, and outputs the generated image.

\subsection{Stage-I: Landmark}
\label{sec:method-landmark}

\begin{figure}[htp]
  \centering
  \includegraphics[width=0.99\linewidth]{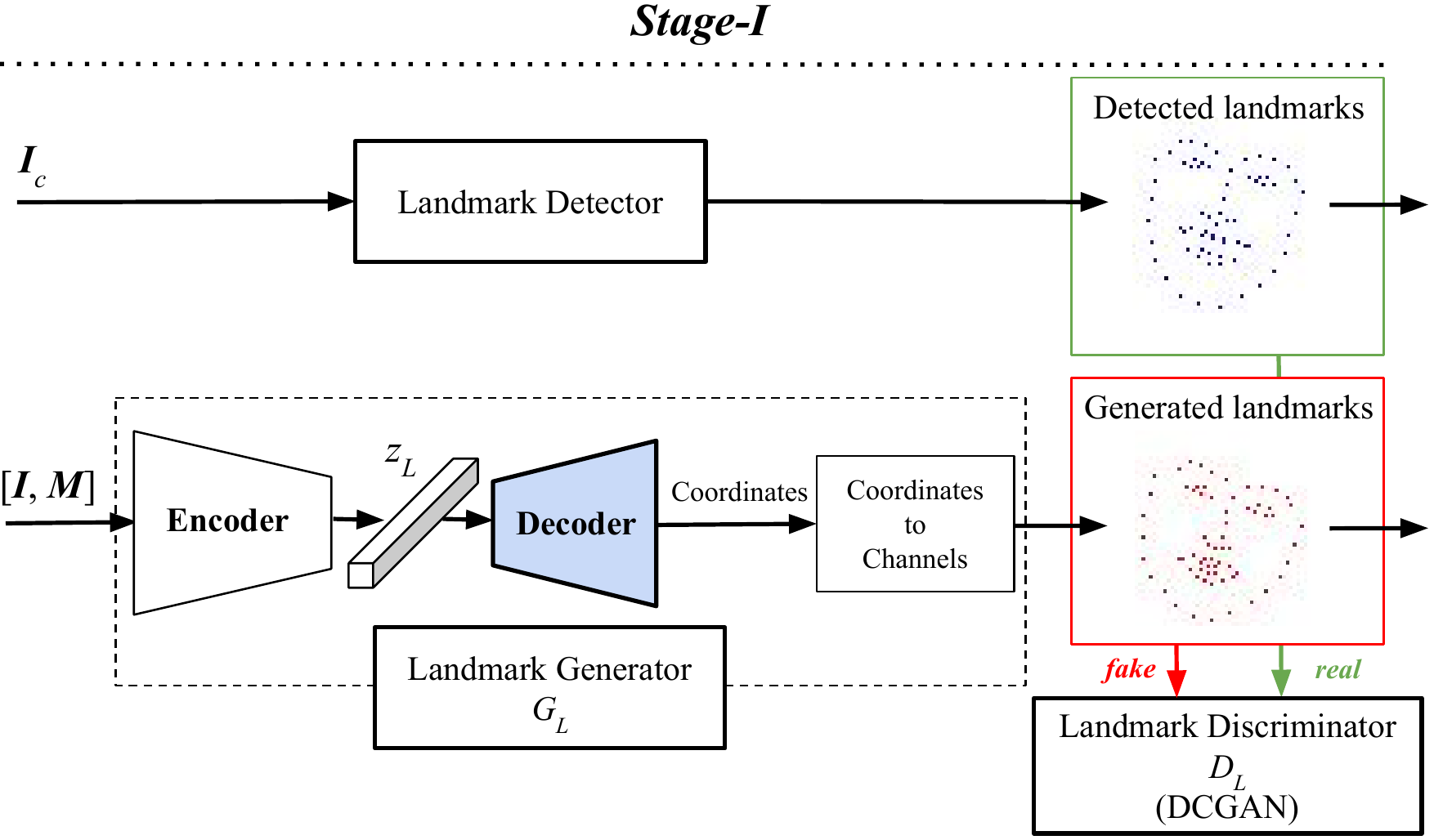}\\
  \caption{Stage-I: Landmark Detection/Generation. The input to detection is the original image $\boldsymbol{I_c}$, and to generation are the blackhead image $\boldsymbol{I}$ and head mask $\boldsymbol{M}$. For the Decoder in $G_L$, we adopt three versions: the decoder from scratch (Scratch); from a pre-trained AE (AEDec); from a pre-trained PDM (PDMDec). } 
\label{framework_stage1}
\end{figure}

The overview of stage-I is shown in Figure~\ref{framework_stage1}. In the case of landmark detection, we simply use the detector implemented in python dlib toolbox~\cite{InterocularDist}. The output are 68 facial keypoints.
In the case of landmark generation, the framework contains an adversarial training by Landmark Generator ($G_L$) and Discriminator ($D_L$).

\myparagraph{Landmark Generator ($G_L$).} $G_L$ has an Auto-encoder structure, and it contains two main parts: Encoder and Decoder. The Encoder compresses the body/scene context of the blackhead image to a latent variable in the bottleneck layer which is then decoded to landmark coordinates by the Decoder. In the following, we describe details of the Encoder and Decoder.

\myparagraph{Encoder of $G_L$.} The input of Encoder are the blackhead image $\boldsymbol{I}$ and a head mask $\boldsymbol{M}$ which indicates the head bounding box. Encoder learns from $\boldsymbol{X} = [\boldsymbol{I}; \boldsymbol{M}]$ to a latent variable $\boldsymbol{z}_L$. The architecture of Encoder has 6 CONV residual blocks, and the latent variable $\boldsymbol{z}_L$ is 32-dimensional.

\myparagraph{Decoder of $G_L$.} Taking $\boldsymbol{z}_L$ as the input, the Decoder works for generating the $2\times 68$ landmark coordinates $\boldsymbol{L}$. Its generic architecture contains 6 FC residual blocks. It is noted that both Encoder and Decoder in $G_L$ are trained from scratch by default.

Training the Encoder and Decoder from scratch 
is challenging in our task, due to diverse body pose and background clutter in social media photos. 
Therefore, we also explore a different route: Instead of simultaneously training both, we first train a \textit{stronger} Decoder and fix it, and then, conditioned on this Decoder, train the Encoder from scratch. Such a procedure is inspired by knowledge transfer between deep models trained on different tasks~\cite{GuptaCVPR16,RussakovskyDSKS15}. 
We implement this by training encorder/decoder pairs with an encoding bottleneck for landmark reconstruction for which the input and output and exactly the same landmark coordinates. During training, the decoder learns the decoding function from the encoding bottleneck to facial landmark coordinates. The pre-trained decoder is then fixed and used as decoder in our landmark generator $G_L$.

Based on different reconstruction methods, we pre-train two types of networks: the classical Auto-encoder (AE) and the landmark-specific Point Distribution Model (PDM)~\cite{PDM}.

\myparagraph{AE decoder (AEDec).} 
The Auto-encoder reconstructs face landmarks using an encoder and a decoder through a bottleneck layer. Both coders are fully connected layers with ReLU activations. $L_2$ loss is used for optimization. 

\myparagraph{PDM decoder (PDMDec).} 
We are using a Point Distribution Model (PDM)  to better represent the 3D pose variations~\cite{PDM,CE-CLM}\footnote{We use the code of \cite{CE-CLM} to train the PDM model~\cite{PDM}. Non-rigid structure from motion~\cite{TorresaniHB03} 
is used to map 2D points to 3D in this code. Our training data are the detected landmarks in PIPA \texttt{TRAIN} set.}. 
Our landmark points are thus parametrized using $\boldsymbol{p}=[s,\boldsymbol{R,t,q}]$ to denote the scale, orientation, translation and non-rigid changes in the following PDM Equation:
\begin{equation}
    \boldsymbol{L}
    = s \cdot \boldsymbol{R} \cdot ({\bar{\boldsymbol{L}}}_{3D} + {\pmb{\Phi}}  \boldsymbol{q}) + \boldsymbol{t}
    \label{eq:pdm}
\end{equation}
where 
${\bar{\boldsymbol{L}}}_{3D}$ denotes the mean value of 3D landmarks mapped from our 2D data, 
${\pmb{\Phi}}$ the $3\times n$ principal component matrix, 
$\boldsymbol{q}$ the $n$-dimensional non-rigid shape parameters, 
$s$ the scaling, 
$\boldsymbol{R}$ are the first two rows of the $3\times 3$ rotation matrix defined by 3 axis angles and $\boldsymbol{t}=[t_x, t_y]$ the translation.
Each landmark vector 
$\boldsymbol{L}$ 
is represented by $n+6$ parameters. 
Equation \ref{eq:pdm} is used as decoder for $G_L$. In the experiments we use $n=34$ principal components to achieve high consistency while allowing for flexibility.

\myparagraph{Loss functions of $G_L$ and $D_L$.} 
We use $L_2$ loss as well as adversarial loss for optimization. 
The adversarial loss is useful because landmarks trained with only $L_2$ loss show noisy alignments, which can be easily detected and remedied by involving a discriminator. 
We adopt the DCGAN discriminator (CONV layers)~\cite{Radford2015-DCGAN} because the landmark channels represent image spatial structures. Landmark coordinates are converted to channels to input to the CONV layers. The convert function is differentiable. Noting that we have also tried a fully-connected discriminator with landmark coordinates as input but it has shown only marginal difference. 

For training $D_L$, any landmark generated by $G_L$ are labeled \textit{fake}, while we use the \emph{detected} landmarks as the \textit{real} examples. Exact losses are formulated as follows:
\begin{align}
    {\cal{L}}_{D_L} =& {\mathbb{E}}_{\boldsymbol{X}  \sim p_{data}(\boldsymbol{X} )}\big[\log{D_L(\boldsymbol{X})}\big] + \nonumber \\  
    &{\mathbb{E}}_{\boldsymbol{X}  \sim p_{data}(\boldsymbol{X} )}\big[\log{(1-D_L(G_L(\boldsymbol{X} )))}\big], 
    \label{eq:land_D_loss} \\
    {\cal{L}}_{G_L} =& {\mathbb{E}}_{\boldsymbol{X}  \sim p_{data}(\boldsymbol{X} )}\big[\log{(D_L(G_L(\boldsymbol{X})))}\big] + \nonumber \\ 
    &{\lambda}_L \| (G_L(\boldsymbol{X} ) - \boldsymbol{L}_d) \|_2, 
    \label{eq:land_G_loss}
\end{align} 
where $\boldsymbol{X}$ is the concatenation of blackhead image $\boldsymbol{I}$ (3 channels) and the head mask $\boldsymbol{M}$ (1 channel). 
$\boldsymbol{L}_d$ is the detected landmark coordinates (\textit{ground truth}). ${\lambda}_L$ is a weight on the $L_2$ loss.

\subsection{Stage-II: Inpainting}
\label{sec:method-inpainting}

\begin{figure}[htp]
  \centering
  \includegraphics[width=0.99\linewidth]{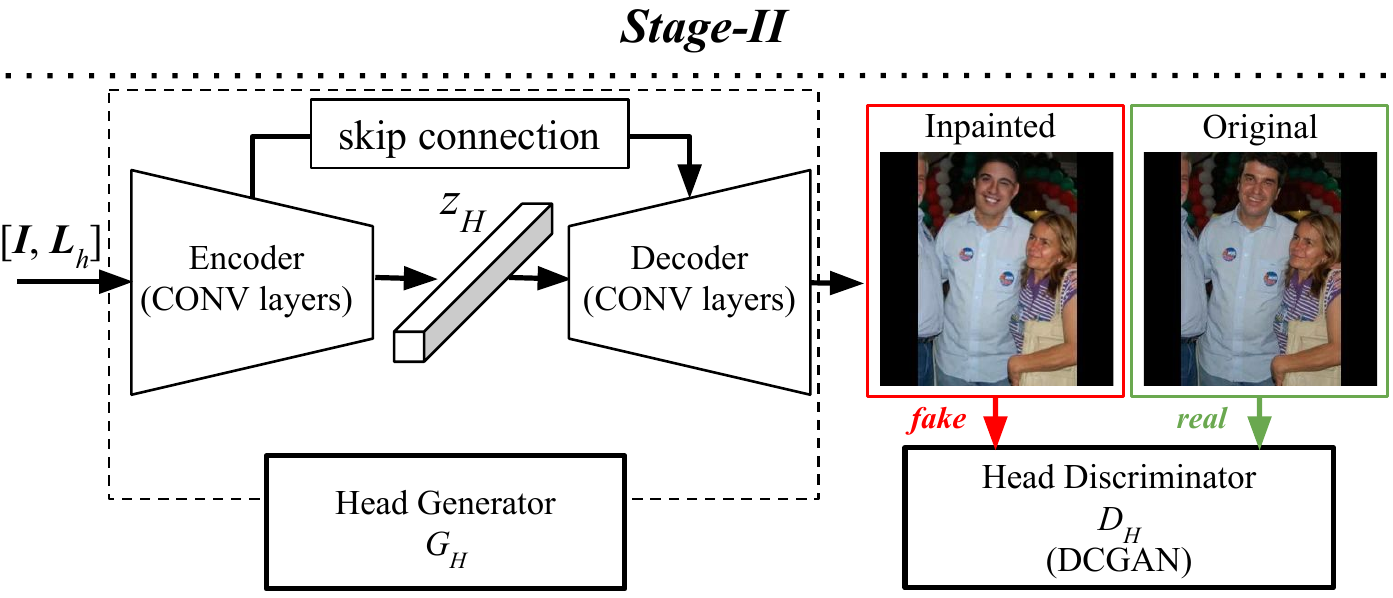}\\
  \caption{Stage-II: Head generation. The input are blackhead image $\boldsymbol{I}$ and landmark channels $\boldsymbol{L}_h$. The generator has an Auto-encoder structure which encodes the input to a bottleneck then decodes to a fake image. The discriminator is the same as in DCGAN~\cite{Radford2015-DCGAN}.}  
\label{framework_stage2}
\end{figure}

Stage-II is conditioned on the landmarks and the blackhead (or blurhead) image to generate head pixels and inpaint the image. As shown in Figure~\ref{framework_stage2}, the architecture is composed of Head Generator $G_H$ and Head Discriminator $D_H$.

\myparagraph{Input.} For head generator $G_H$, the 68 landmark heatmaps $\boldsymbol{L}_h$ are concatenated with the head-obfuscated image $\boldsymbol{I}$ as input.
As the head region is obfuscated, the landmark keypoints guide the pixel generation.

For the input of the head discriminator $D_H$, we fuse the generated head image with the black head image as \textit{fake} and treat the original image as \textit{real}. Then, we feed the \textit{fake}, \textit{real} pairs into the head discriminator. It is worth noting that we use the whole body image instead of head regions in order to generate a realistic image which has natural transition between head and surroundings including body and background.

\myparagraph{Head Generator ($G_H$) and Discriminator ($D_H$).} The head generator $G_H$ is a ``U-Net''-based architecture~\cite{U-net}, \ie, convolutional Auto-encoder with skip connections between encoder and decoder to help propagate image information directly from input to output. It generates a natural head image according to both surrounding context and landmarks. 
The architecture of the head discriminator is the same as in DCGAN~\cite{Radford2015-DCGAN}.

\myparagraph{Loss function.} We use both $L_1$ loss and adversarial loss to optimize $G_H$ and $D_H$:
\begin{align}
    {\cal{L}}_{D_H} =& {\mathbb{E}}_{\boldsymbol{Y} \sim p_{data}(\boldsymbol{Y})}\big[\log{D_H(\boldsymbol{Y})}\big] + \nonumber \\  
    &{\mathbb{E}}_{x \sim p_{data}(\boldsymbol{Y})}\big[\log{(1-D_H(G_H(\boldsymbol{Y})))}\big], 
    \label{eq:head_D_loss} \\
    {\cal{L}}_{G_H} =& {\mathbb{E}}_{\boldsymbol{Y} \sim p_{data}(\boldsymbol{Y})}\big[\log{(D_H(G_H(\boldsymbol{Y})))}\big] + \nonumber \\ 
    &{\lambda}_H \| (G_H(\boldsymbol{Y}) - \boldsymbol{I}_{c}) \|_1, 
    \label{eq:head_G_loss}
\end{align} 
where  $\boldsymbol{Y}$ is the concatenation of blackhead image $\boldsymbol{I}$ and the landmark heatmaps $\boldsymbol{L}_h$ mapped from landmark coordinates. $\boldsymbol{I}_{c}$ is the original image (\textit{ground truth}). ${\lambda}_H$ is the weight of $L_1$ loss\footnote{Detail architecture and hyper-parameters are given in the supplementary materials.}.

\section{Experiments}
\label{sec:experiments}
We evaluate the presented two-stage head inpainting pipeline on a social media dataset in terms of inpainting appearance and pose plausibility, as well as identity obfuscation performance against machine recognizers. We analyze the impact of different input types (original, blackhead, and blurhead) and different choices of decoders and losses for landmark generation (\S\ref{sec:method-landmark}).

\subsection{Dataset}

Since we need to evaluate our method on realistic social media photos, we use the PIPA dataset~\cite{PIPA}. It is the largest social media dataset to date (37,107 Flickr images with 2,356 annotated individuals), and contains people in diverse events, activities, and poses. Each of 63,188 person instances are annotated with head bounding boxes. 
 
In order to maximize the amount of training data, we have introduced new training-test splits over PIPA, instead of resorting to existing ones. We split 2,356 PIPA identities into \texttt{TRAIN} set (2,099 identities, 46,576 instances) and \texttt{TEST} set (257 identities, 5,175 instances). We have further pruned both sets with heavy profile or back-view heads, resulting in 34,383 instances in \texttt{TRAIN} and 1,909 in \texttt{TEST}. The \texttt{TRAIN} set is used for training landmark and head generators. \texttt{TEST} set is the evaluation set. 

Our landmark and inpainting generators take a fixed-size input ($256\times 256\times 3$). For every training and testing sample, we prepare the input by first obtaining the \textit{body crop}. We follow the procedure in \cite{Oh2015Iccv}: extend the head box with fixed ratios ($3\times$width and $6\times$height),
and then resize and zero-pad the body crop such that it fits tightly in the square $256\times 256$.

\subsection{Scenarios and inputs}
Our approach introduced in \S\ref{sec:method} is versatile and supports scenarios where the user (who wants to obfuscate an image) has access to the original image or only has access to already head-obfuscated images (e.g. blacked out). The necessity for this versatility is 
that social network service providers may aim to upgrade the privacy level by obfuscating images through blurring or blacking-out heads, even though it has been shown to be quite ineffective~\cite{joon16eccv}. 

In order to simulate multiple scenarios, we consider three types of inputs to our obfuscator: original, blackhead, or blurhead, where the latter two are common obfuscation techniques these days. We prepare blackhead and blurhead inputs following the procedure in~\cite{joon16eccv}. PIPA head box annotations indicate the head region to be obfuscated, which is either filled in with black pixels or smoothed with a Gaussian blur kernel  specified in \cite{joon16eccv}.

\begin{table*}
\centering
\small
\caption{Evaluation of proposed obfuscation methods. We quantify the quality of the proposed obfuscation method against landmark quality, inpainting quality, as well as obfuscation effectiveness (person recognition rates). We vary the loss ($D_L$ here represents the adversarial loss) and decoder used in our landmark generator (\S\ref{sec:method-landmark}); the head inpainter is always the $G_H$ + $D_H$(\S\ref{sec:method-inpainting}).
}
\begin{centering}
\begin{tabular*}{17cm}
{ l l l c c c c c c c c c c}
\multicolumn{3}{c}{Obfuscation method}&\multicolumn{10}{c}{Evaluation}\\
\cmidrule{1-3}\cmidrule{5-13}
&\multicolumn{2}{c}{Landmark}&& \multicolumn{2}{c}{Landmark} && \multicolumn{2}{c}{Inpainting} && \multicolumn{3}{c}{Person recognizer}\\
\cmidrule{2-3}\cmidrule{5-6}\cmidrule{8-9}\cmidrule{11-13}
Input &Loss&Decoder  && $L_2$ & Norm. $L_2$ && SSIM & mask-SSIM && \texttt{head} & \texttt{body+head} & head contrib.\\
\cmidrule{1-3}\cmidrule{5-6}\cmidrule{8-9}\cmidrule{11-13}
Original & \multicolumn{2}{l}{No head inpainting}&& / & / && 1.000 & 1.000 && 85.6\% & 88.3\% & 72.2\% \\
Original&\multicolumn{2}{l}{NN head copy-paste} && / & / && 0.872 & 0.195 && 1.2\% & 7.1\% & 67.5\% \\
\cmidrule{1-3}\cmidrule{5-6}\cmidrule{8-9}\cmidrule{11-13}
Blur&\multicolumn{2}{l}{No head inpainting} && / & / && 0.931 & 0.396 && 52.2\% & 71.6\% & 3.2\% \\
Blur&\multicolumn{2}{l}{Detected landmarks} && 0.00 & 0.000 && 0.962 & 0.679 && 43.7\% & 51.7\% & 70.8\% \\
Blur&$L_2$&Scratch && 6.32 & 0.230 && 0.954 & 0.578 && 36.2\% & 48.4\% & 66.8\% \\
Blur&$L_2$+$D_L$&Scratch && 4.85 & 0.182 && 0.955 & 0.586 && 38.0\% & 48.4\% & 66.6\% \\
Blur&$L_2$+$D_L$&AEDec && 4.77 & 0.180 && 0.951 & 0.585 && 37.5\% & 48.0\% & 66.1\%  \\
Blur&$L_2$+$D_L$&PDMDec && 4.50 & 0.168 && 0.953 & 0.593 && 37.9\% & 49.1\% & 66.7\%  \\
\cmidrule{1-3}\cmidrule{5-6}\cmidrule{8-9}\cmidrule{11-13}
Black&\multicolumn{2}{l}{No head inpainting} && / & / && 0.000 & 0.000 && 2.1\% & 67.0\% & 14.0\% \\
Black&\multicolumn{2}{l}{Detected landmarks} && 0.00 & 0.000 && 0.902 & 0.405 && 10.1\% & 21.4\%  & 70.8\% \\
Black&\multicolumn{2}{l}{NN landmarks} && 2.48 & 0.088 && 0.896 & 0.332 && 7.9\% & 20.4\% & 71.3\% \\
Black&$L_2$&Scratch && 13.6 & 0.501 && 0.884 & 0.186 && 5.8\% & 17.4\% & 73.6\% \\
Black&$L_2$+$D_L$ &Scratch && 13.0 & 0.477 && 0.882 & 0.191 && 5.8\% & 17.2\% & 71.4\% \\
Black&$L_2$+$D_L$ &AEDec && 11.7 & 0.431 && 0.885 & 0.199 && 5.6\% & 17.4\% & 72.5\% \\
Black&$L_2$+$D_L$ &PDMDec && 12.3 & 0.453 && 0.885 & 0.196 && 5.6\% & 17.4\% & 71.0\%  \\
\cmidrule{1-3}\cmidrule{5-6}\cmidrule{8-9}\cmidrule{11-13}
\end{tabular*}
\end{centering}
\vspace{-1mm}
\label{Table_1_quantitative}
\end{table*}

\subsection{Quantitative results}

Here, we quantify the intermediate face landmark quality as well as the obfuscation performance of the final head-inpainted images. 

\subsubsection{Landmark}
\label{exp:landmark_model_evaluation}

Facial landmarks may be either detected or generated depending on the input type (head visible or not; see \S\ref{sec:method-landmark}). In this section, we evaluate the generated landmark quality
in terms of the $L_2$ distance to the detected landmarks (used as \textit{ground truth}) and the $L_2$ distance normalized by the inter-ocular distance~\cite{InterocularDist}. Although detected landmarks are not perfect, they turned out to be good proxies to the ground-truth in our preliminary inspection on real face images.

We investigate three axes of factors for our landmark generator. (1) the input type: original, blackhead, or blurhead. (2) the loss function: only $L_2$ versus $L_2$ and adversarial loss ($D_L$). (3) the decoder type: trained from scratch, autoencoder pretrained (AEDec), or Point Distribution Model pretrained (PDMDec). A summary of the quantitative results is given in Table \ref{Table_1_quantitative} (``Landmark'' column). Note that for the original images, our best landmark generator achieves an $L_2$ distance of 2.41 on average (not shown in table) -- which gives an upper bound (best-case) on the generated landmark quality. 

\myparagraph{Input type.}
We compare the $L_2$ distance between generated and detected landmarks for three types of inputs: original, blackhead, or blurhead. For original images, we use detected landmarks, which gives by definition zero $L_2$ distance. We observe from Table \ref{Table_1_quantitative} that head-blurred inputs show consistently closer landmark locations to the detected landmarks: e.g. 6.32 versus 13.6 for landmark generator with decoder trained from scratch with $L_2$ loss. Blurhead images indeed contain structural information about the face keypoints.

\myparagraph{Loss function.}
We compare two choices of loss functions: only $L_2$ versus $L_2+D_L$. Given a blackhead input with landmark decoder trained from scratch, using only $L_2$ loss yields the 13.6 distance from the detected landmarks. Adding adversarial loss $D_L$ marginally improves the distance to 13.0. However, for head-blurred images, the improvement due to adversarial loss is much greater (from 6.32 to 4.85).

\myparagraph{Decoder.}
We consider three choices of decoder in the landmark generator $G_L$: learning from scratch (Scratch), pre-trained with AE (AEDec), and pre-trained with PDM (PDMDec). For both blurhead or blackhead cases, conditioning the decoder with either AEDec or PDMDec helps generating landmarks closer to the detected ones: e.g. for blackhead input, $L_2$ distance metric improved from 13.0 to 11.7 and 12.3, respectively, although the impact is less dramatic than for the type of input. 

\subsubsection{Inpainting}
We evaluate the head generation quality comparing to original images (\textit{ground truth}) using SSIM~\cite{ImageQuality} for whole image and mask-SSIM~\cite{PG2} for head region only.
One simple baseline of inpainting is using Nearest Neighbor (NN) head\footnote{NN head is searched in training data based on the mean $L_2$ distance of detected landmarks.} to do copy-paste. This ignores the blending with surroundings, resulting in unpleasant visualization and a quite low SSIM score of 0.872, lower than all our inpainting results.

\subsubsection{Obfuscation}
While it is interesting to see how well facial landmarks can be hypothesize even images where the face is blurred or blacked-out, a more important question is how well our inpainting methods can fool automatic person recognizers. 
For this, we have taken person recognizer models from \cite{Oh2015Iccv}, retrained them on our data-splits 
to measure the obfuscation performance, in terms of the drop in recognition rate compared to the non-obfuscated case. We also provide a rationale for our good obfuscation performance based on the analysis of the attention of the recognizer. 

\myparagraph{Person recognizer.}
We use the social media person recognition framework \texttt{naeil} \cite{Oh2015Iccv}. Unlike typical face recognizers, \texttt{naeil} uses also body and scene context cues for recognition. It has thus proved to be relatively immune to common obfuscation techniques like blacking or blurring head regions~\cite{joon16eccv}.

Following \cite{Oh2015Iccv}, we first train feature extractors over head and body regions, and then train an SVM identity classifier on top of those features. We may also concatenate features from multiple regions (e.g. head+body) to allow it to extract cues from multiple regions. In our work, we use GoogleNet features from \texttt{head} and \texttt{head+body} for evaluation of obfuscation performance.

Although we present results against a particular instance of machine recognition system, the obfuscation method is \emph{target-generic}: the head generation is not conditioned on any particular recognition system. It is thus expected to work against a generic machine recogniser. We have also verified that the obfuscation results show similar trends against AlexNet-based analogues (supplementary materials).

\myparagraph{Head inpainting provides good protection.}
Table \ref{Table_1_quantitative} shows  obfuscation performance (columns \texttt{head} and \texttt{head+body}). Under no obfuscation, the \texttt{head+body} recognition performance is 88.3\%. Black/blurring baselines give 67.0\%, and 71.6\%, respectively -- confirming the observation in \cite{joon16eccv} that these are ineffective. 
On the other hand, our head inpainting methods show $<50\%$ (blurhead input) and $<21\%$ (blackhead input) recognition rates for \texttt{head+body} recognizers. They are more effective protection techniques than blacking or blurring head regions.

\myparagraph{Cues used.}
We compare the recognition rates between \texttt{head} and \texttt{head+body}. When the recognizer relies solely on head cues, while the head has been inpainted, then the recognition rates are lower than the \texttt{head+body} counterparts. For example, the last row method against \texttt{head} recognizer gives $5.6\%$ versus $17.4\%$ for \texttt{head+body}, nearly reaching the chance level recognition rate ($2.1\%$). 

\myparagraph{Input type.}
While having access to blurred head images help generating more plausible landmarks (\S\ref{exp:landmark_model_evaluation}) as well as visually natural head inpainting (\S\ref{sec:qualitative}), they may leak identity information. We compare the recognition rates when either blurhead or blackhead inputs are used. Our head inpainting based on blackhead result in $17\%\sim21\%$ accuracy, while blurhead based results are in the range $48\% \sim 50\%$ accuracy. This confirms that indeed there exists a trade-off between plausibility of generated heads and the obfuscation performance.

\myparagraph{Detected vs generated landmarks.}
While identity information may leak through blurred heads, it may also leak through the landmark detections (face shape). On the other hand, generated landmarks enjoys the possibility to come up with an equally plausible landmark hypothesis but with different face shapes. For the blackhead input, the detected landmarks indeed result in higher recognition rate ($21.4\%$) than generated ones (e.g. $17.4\%$ on last row), with similar trend for the blurhead cases. 

\myparagraph{Rationale for good obfuscation -- recognizer attention.}
We have verified that our head obfuscation scheme exhibits better performance than commonly used ones like blacking and blurring. We give a rationale for this phenomenon by means of the \emph{recognizer attention}. Given an input, \emph{recognizer attention} refers to the regions in the image where the recognizer extracts cues from. We hypothesize that while blacked or blurred heads induce recognizer attention on non-head regions, our inpainted heads attract attention on the heads.

For the  \emph{recognizer attention} we have used the gradient-based mechanism from Simonyan \etal~\cite{simonyan14iclrw}. We first compute the gradient of the neural network prediction with respect to the input image; take maximal absolute values along the RGB channel; and then smooth with Gaussian blurring. To quantify the chance of attending on the head region, we have computed the ``head contribution'' score by estimating 
\begin{align*}
\text{head contrib.}=\mathbb{P}[\text{max attention is inside head region}]
\end{align*}
over the test samples.

See final column of table~\ref{Table_1_quantitative} for the results. We observe that while the original image has $72.2\%$ chance of inducing attention on the head region, blacked or blurred heads are much less likely to attract the recognizer's attention ($14.0\%$ and $3.2\%$, respectively). This explains why \texttt{head+body} is still performing well: it simply ignores the confusing head cue. On the other hand, our inpainting-based obfuscation still attracts the recognizer's attention as much as the non-obfuscated head image does ($71.0\%$ versus $72.2\%$). This indicates that the realism of inpainted heads encourages the recognizer to still rely its decision on the inpainted head, effectively leading to misjudgment by the recognizer.

\begin{figure*}[htp]
  \centering
  \includegraphics[width=0.99\linewidth]{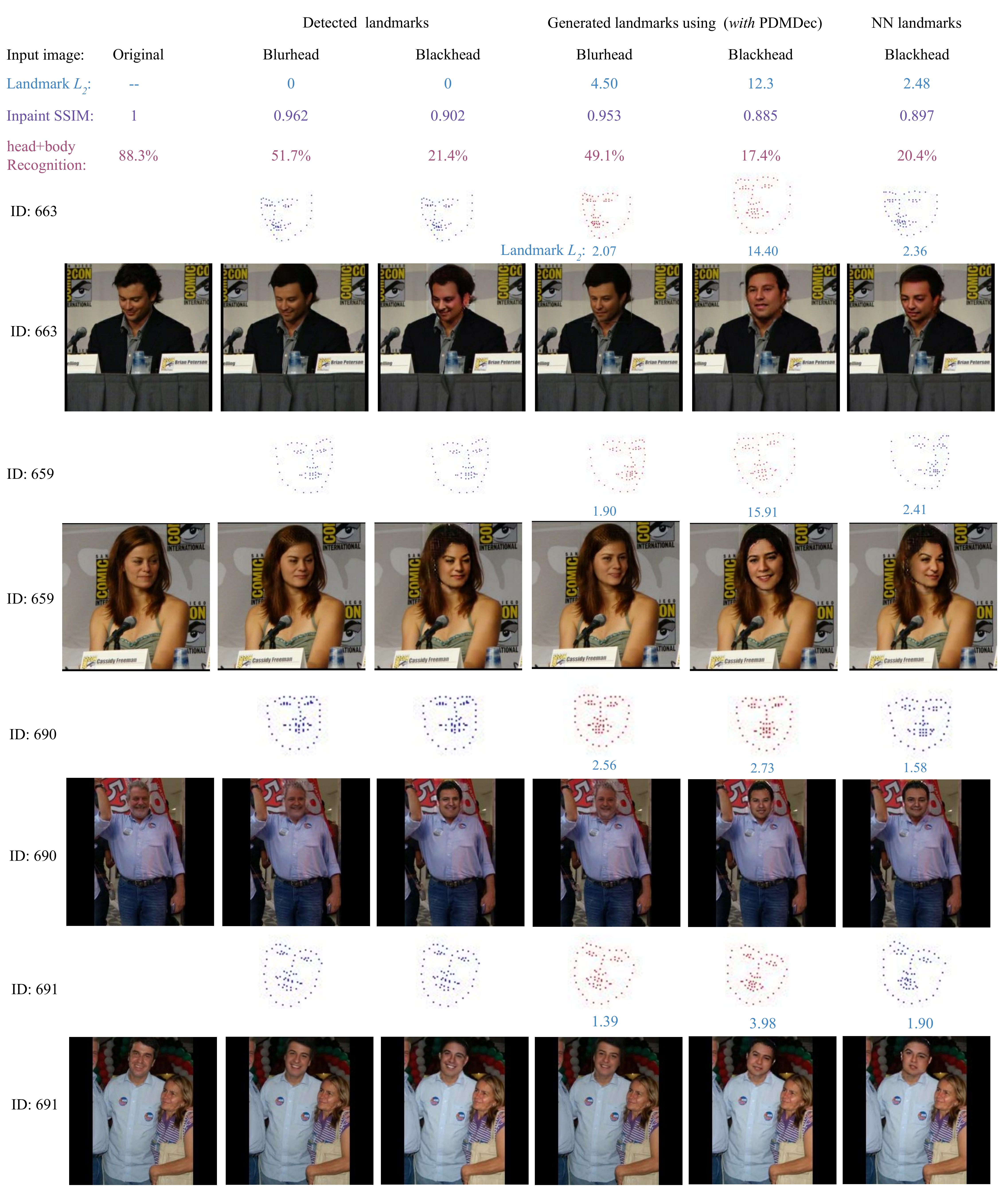}\\
   \vspace{-0.1cm}
     \caption{Visualization results on PIPA dataset. We show the head inpainting results using detected and generated landmarks (from the PDMDec model). Top rows present key quantitative numbers for reference. Landmark generation error (distance to the detected one) is also given for each single instance (under landmark image).}
  \label{landmark_visualization}
\end{figure*}

\subsection{Qualitative results}
\label{sec:qualitative}
Obfuscation patterns should not only be effective against recognizers, but also look natural for the applicability in social media. In this section, we visually examine the generated landmarks and corresponding inpainted heads.

For generating natural heads, landmarks should look like that of an actual face and be consistent with the body pose. However, at the same time obfuscation performance benefits from landmarks that do not preserve the original face shape. In this section, we discuss if our generated landmarks achieve both realism, while effectively obfuscating machine recognizers. Qualitative results are given in Figure \ref{landmark_visualization}.

\myparagraph{Detected versus generated landmarks.}
Given an original image with a visible head, we detect landmarks, while for blackhead we hypothesize them from regions other than the head itself. The comparison between columns 2,3 (detected landmarks) and columns 4,5 (generated landmarks) in Figure \ref{landmark_visualization} illustrates the difference. In all the examples shown, the detected landmarks closely follow the original image. On the other hand, the generated landmarks, especially for blackhead cases, results in landmarks and head inpainting with different head poses. However, the generated landmarks are still plausible with respect to the body pose and activity. Finally, note that by generating landmarks, we can further mask identity information (recognition rates are consistently lower for inpainting based on generated landmarks), while keeping reasonable realism.

\myparagraph{Blackhead versus blurhead.}
Landmarks may be generated from either blurhead or blackhead images. We visualize how the head information contained in blurred cases improve the inpainting quality. Columns 2,4 and columns 3,5 in Figure~\ref{landmark_visualization} show respective examples for blur and black cases. Involving blurred head images during landmark and head generation results in inpainting that resembles the original head, especially the head pose and hair color/style (\eg ID-690). On the other hand, not providing any information in the head region results in a significantly different, yet plausible, head images. In particular, when even landmarks are generated, the resulting head images are drastically different from the original one. Such a shift of appearance is reflected in the low recognition rate ($17.4\%$).

\begin{table}[hbp]
\centering
\footnotesize
\caption{Human perceptual study (HPS) scores and landmark detection success ratios (LDSR). Landmarks are from ``detected'', and ``generated'' by PDMDec methods.}
\begin{tabular*}{8cm}
{@{\extracolsep{\fill}} m{3em} @{\extracolsep{\fill}} m{3em} @{\extracolsep{\fill}} m{3em} m{3em} m{3em}  m{3em} m{3em}}
\toprule 
& & & \multicolumn{2}{c}{blurhead(Ours)} & \multicolumn{2}{c}{blackhead(Ours)} \\
\cmidrule{4-5}  \cmidrule{6-7}
& Orig. & CE~\cite{contextEncoder} & detected & generated & detected & generated \\
\midrule[1pt]	
HPS: &	0.93 & 0.04 & 0.60 & 0.39 & 0.19 & 0.11 \\
\midrule[1pt]	
LDSR: &	1.00 & 0.36 & 1.00 & 0.95 & 0.99 & 1.00 \\
\bottomrule[1pt]
\end{tabular*}
\label{tab:userstudy}
\vspace{-0.2cm}
\end{table}

\subsection{Comparing with the state-of-the-art}
We considered two related works for inpainting~\cite{contextEncoder, brkicCVPRW17}. We have implemented the Context Encoder (CE)~\cite{contextEncoder} on the PIPA dataset, but skipped the visual result as the image quality was not competitive. We include it by the score of human perceptual study (HPS) in Table~\ref{tab:userstudy}. 
Another related work~\cite{brkicCVPRW17} focuses on full body replacement using body contours. 
Again, their reported visualization results are far from being competitive especially on head regions. 

We perform the HPS on Amazon Mechanical Turk (AMT). For each method, we show 55 real and 55 inpainted images in a random order for 20 users. Users press the real or fake button for an image within 1s and the first 10 images are only practices~\cite{PG2,IsolaZZE17}.
The first row of Table~\ref{tab:userstudy} contains the ratios of images that were judged as real for different methods: (1) original unaltered; (2) inpainted by Context Encoder (CE)~\cite{contextEncoder} (blackhead image as input); (3) inpainted by our four models.
We observe that (1) assessment on original unaltered images is $93\%$ real -- not perfect (2) using same blackhead images with additional landmarks generated by our model, we can confuse $11\%$ of the users -- nearly threefold increase w.r.t. CE baseline (3) we achieve an 8pp improvement when using detected landmarks (4) we get significant higher fooling rates ($60\%$, $39\%$) in the blurhead cases. So while our fooling rate is not perfect these numbers are encouraging and improve over related works.

For full comparisons, we also measure the landmark detection success ratio (LDSR) inspired by~\cite{IsolaZZE17}. We use the landmark detector for head-inpainted images and recored the success detection ratios.
Intuitively, LDSR should be higher for heads inpainted with better synthesis models. As shown in Table~\ref{tab:userstudy}, heads inpainted by our methods have LDSR above 95\%, while CE has only 36\% - our methods generate heads with much clearer face structures.


\section{Conclusion}
\label{sec:conclusions}

To address the problem of obfuscating identities in social media photos, we have presented a two-stage head inpainting method. Although the social media setup is more challenging than previous face-generation setups (diverse head and body poses and backgrounds), our method has proved to generate both natural and effective obfuscation patterns that effectively confuses an automatic person recognizer. In particular, our method is \emph{target-generic}: the obfuscation is designed to work against any recognizer, be it human or machine. Also, the method does not require access to the original image, enabling to ``upgrade'' existing obfuscation patterns to our privacy-enhanced version.

\section*{Acknowledgments}
This research was supported in part by German Research Foundation (DFG CRC 1223),
Toyota Motors Europe.

{\small
\bibliographystyle{ieee}
\bibliography{egbib}
}

\clearpage

\setcounter{section}{0}
\renewcommand\thesection{\Alph{section}}
\noindent
{\Large {\textbf{Supplementary materials}}}
\\

These supplementary materials include additional details in network architecture (\S\ref{suppsec:arch}) and training (\S\ref{suppsec:implementation}), as well as extended figures and tables: \S\ref{suppsec:vis} and \S\ref{suppsec:alexnet} introduce extensions of Figure 5 and Table 1 in the main paper, respectively.


\section{Network architectures}
\label{suppsec:arch}
\begin{figure*}[htp]
  \centering
  \includegraphics[width=0.9\linewidth]{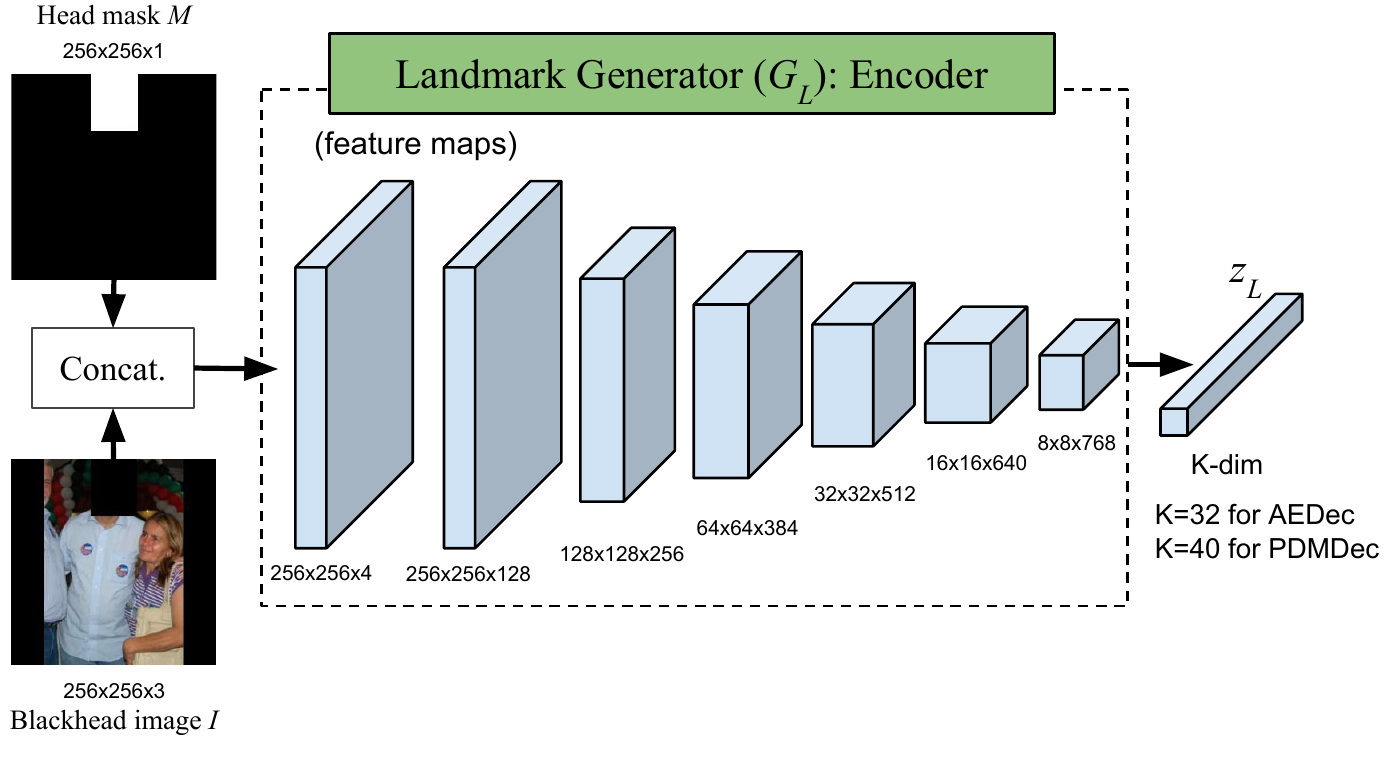}\\
   \vspace{-0.1cm}
     \caption{The architecture of the Encoder used in Landmark Generator $G_L$.}
  \label{supp_arch_land_G_encoder}
\end{figure*}

\begin{figure*}[htp]
  \centering
  \includegraphics[width=0.99\linewidth]{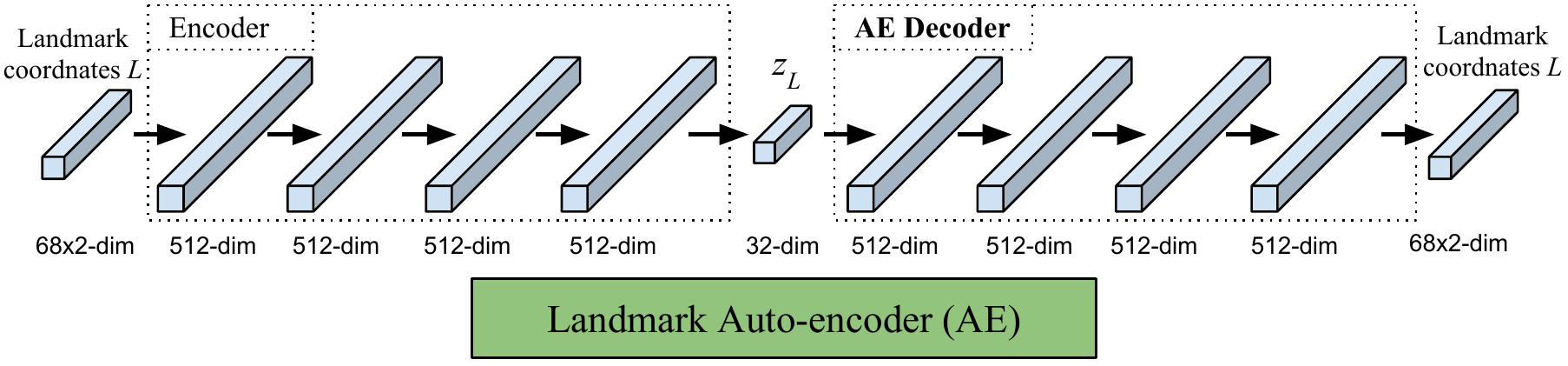}\\
   \vspace{-0.1cm}
     \caption{The architecture of the Auto-encoder used for pre-training the AE Decoder (AEDec). The pre-trained AE Decoder will be connected to $G_L$ Encoder through the bottleneck layer $\boldsymbol{z}_L$.}
  \label{supp_arch_land_ae}
\end{figure*}

\begin{figure*}[htp]
  \centering
  \includegraphics[width=1\linewidth]{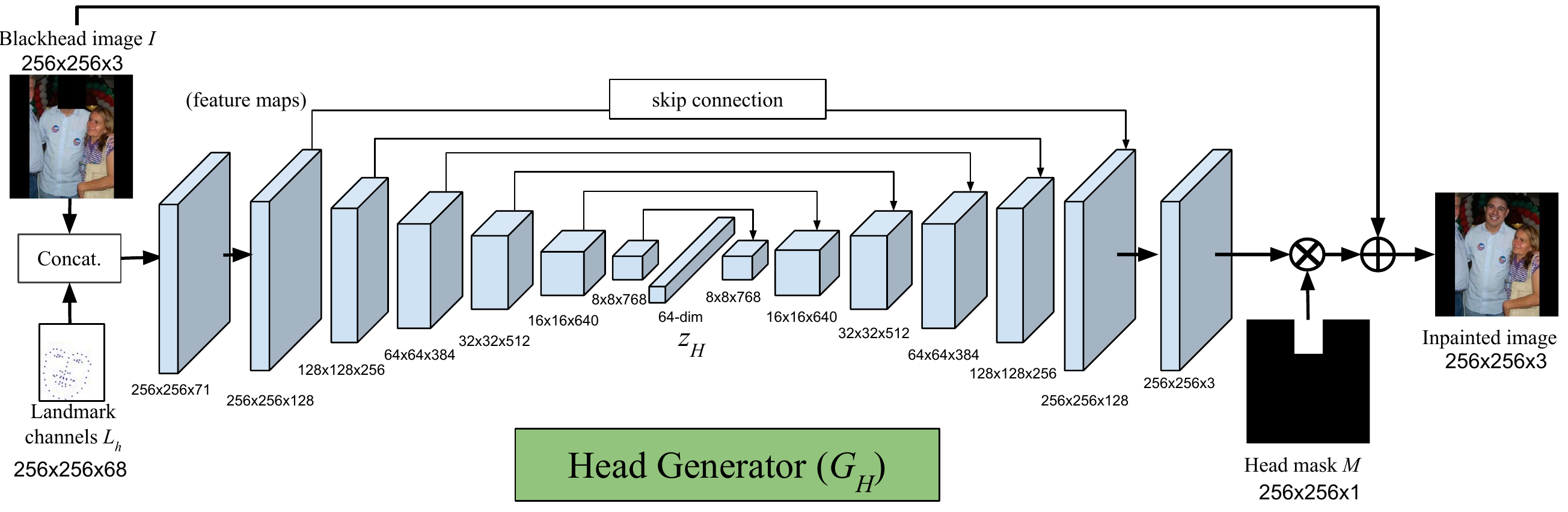}\\
   \vspace{-0.1cm}
     \caption{The network architecture of Head Generator $G_H$. It is based on the ``U-net''. Noting that the landmark channels are 256x256x68 but are cropped to the head region size in this figure only for a better visualization of points.}
  \label{supp_arch_head_G}
\end{figure*}

In Figure~\ref{supp_arch_land_G_encoder}, Figure~\ref{supp_arch_land_ae} and Figure~\ref{supp_arch_head_G}, we present three architectures respectively for the Encoder of Landmark Generator $G_L$, the landmark Auto-encoder (for pre-training the AEDec) and the Head Generator $G_H$. 

In Figure~\ref{supp_arch_head_G}, we should note that the output of the deep network is the intact image (256x256x3) including the body and head. It is then post-processed by cropping and pasting based on the head mask and the blackhead image. Therefore, in the final output only the head region is generated.

\section{Implementation details}
\label{suppsec:implementation}
Both the landmark generator and head generator are trained with the Adam optimizer ~\cite{Adam} with the weights $\lambda_L = 2$ (in the main paper Equation (3)) and $\lambda_H = 50$ (in the main paper Equation (5)). Initial learning rates (for both generator and discriminator) are $2\times 10^{-5}$, and it decays to half every $5,000$ iterations.

For landmark generation models, 
the minibatch size of landmark generation models is set to $16$; optimization stops after $10,000$ iterations; each iteration consists of $5$ and $1$ parameter updates for the generator and the discriminator, respectively. We have $34,383$ training data in total. Therefore, it is about $23.3$ epoches for training the generator and $4.7$ epoches for training the discriminator.
For AEDec, we train the landmark Auto-encoder with the minibatch size $16$; optimization stops after $60,000$ iterations.

For head generation models, 
the minibatch size of landmark generation models is set to $6$; optimization stops after $13,000$ iterations; each iteration consists of $5$ and $1$ parameter updates for the generator and the discriminator, respectively. It is about $8.7$ epoches for training the generator and $1.7$ epoches for training the discriminator.

\section{Visualization results}
\label{suppsec:vis}

In this section, we show the visualization results using different landmark generation models, as a supplement to the Figure 5 of main paper. Specific landmark models are $L_2$
($L_2$ loss was used in the Table 1 of main paper) with Scratch Decoder, $L_2+D_L$ with Scratch Decoder, $L_2+D_L$ with AE Decoder and $L_2+D_L$ with PDM Decoder. 

\begin{figure*}[htp]
  \centering
  \includegraphics[width=0.99\linewidth]{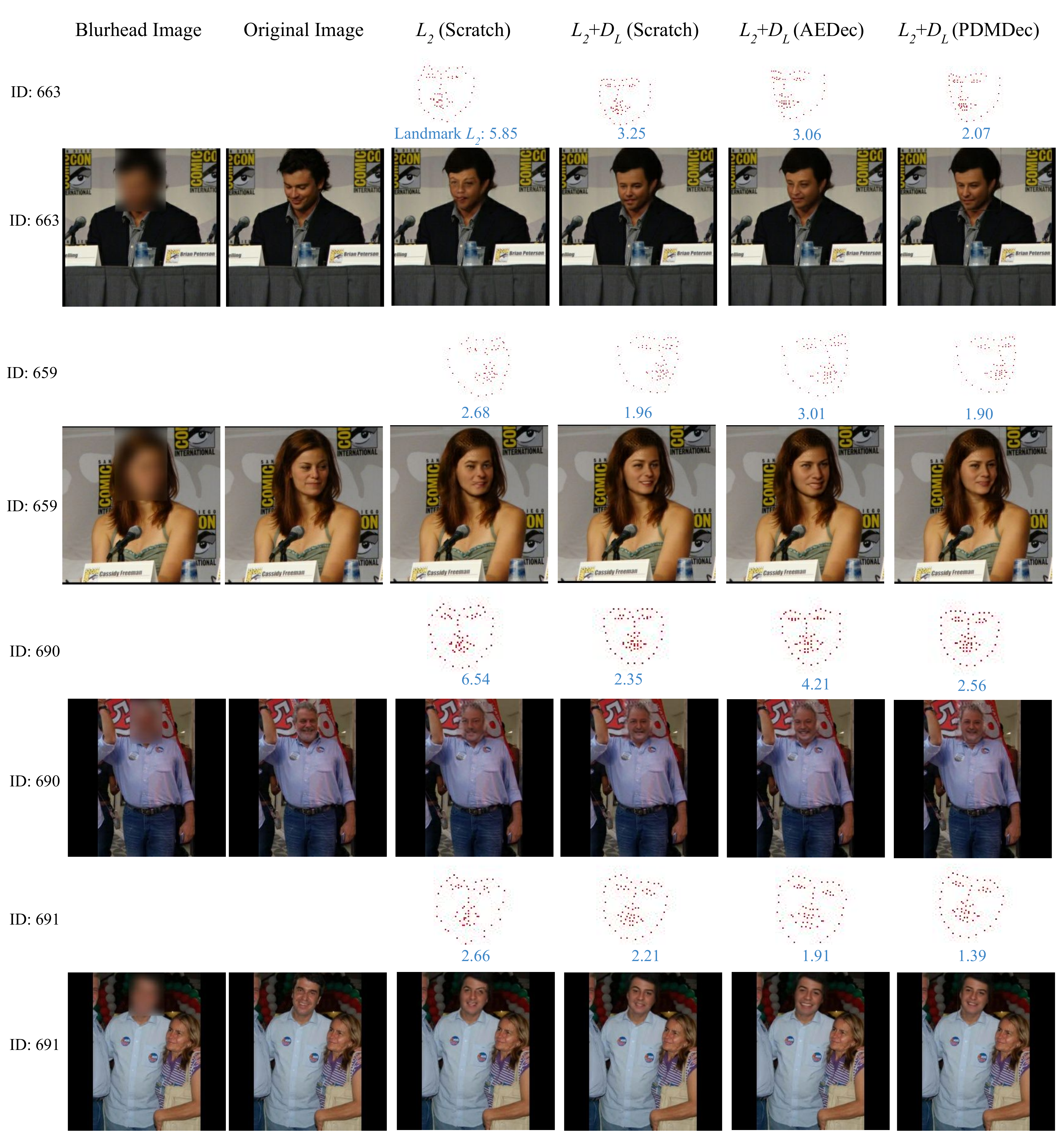}\\
   \vspace{-0.1cm}
     \caption{Visualization results on PIPA dataset. The input is blurhead image both for landmark generation and head generation. Landmark generation error (the distance to the detected ones) is given under each instance.}
  \label{supple_Result_Figure_Landmark_blur}
\end{figure*}

Figure~\ref{supple_Result_Figure_Landmark_blur} presents the results with blurhead images as input. In most cases, we achieve the best visual quality as well as the lowest landmark generation errors using the PDMDec model. 

\begin{figure*}[htp]
  \centering
  \includegraphics[width=0.99\linewidth]{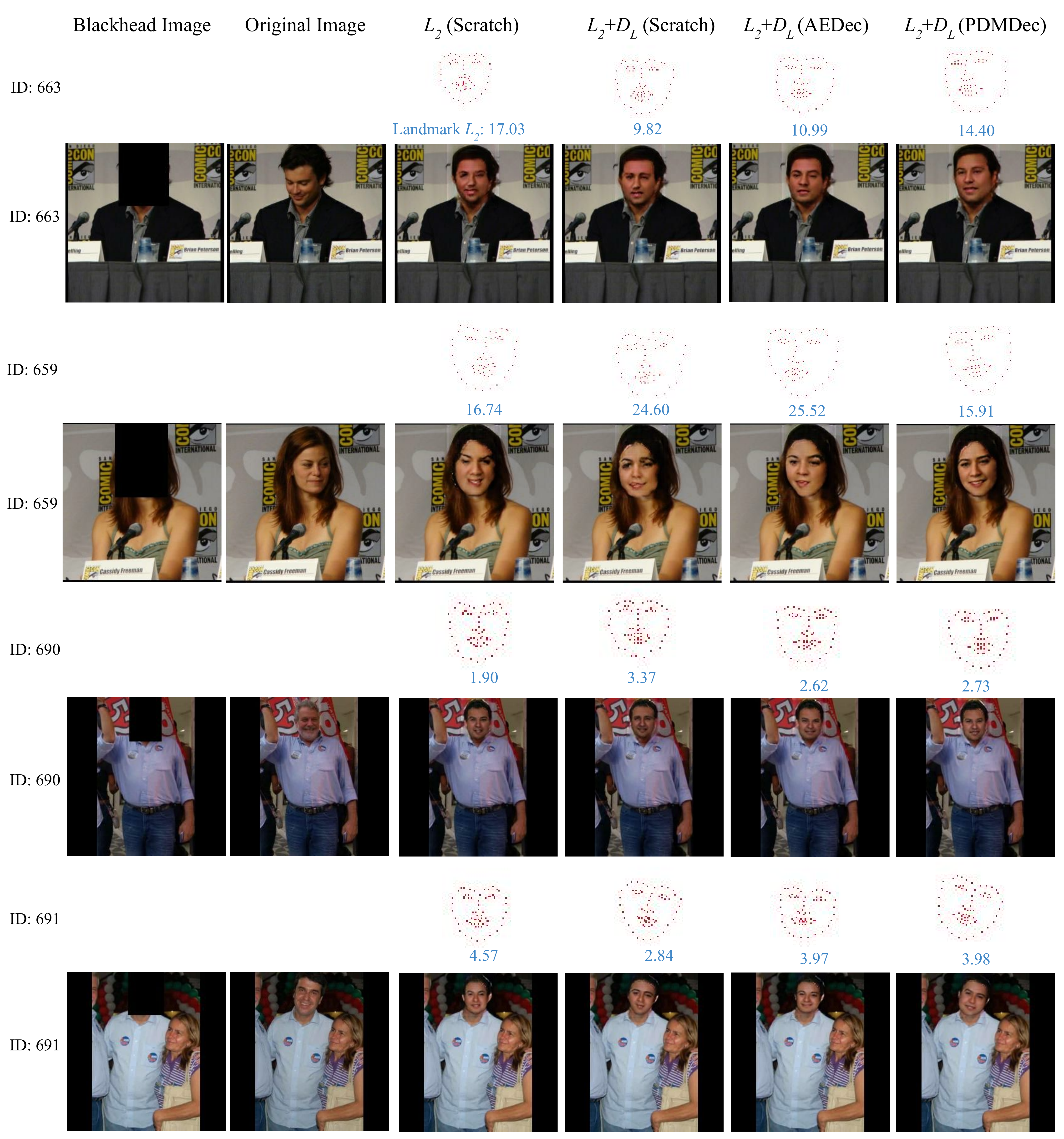}\\
   \vspace{-0.1cm}
     \caption{Visualization results on PIPA dataset. The input is blackhead image both for landmark generation and head generation. Landmark generation error (the distance to the detected ones) is given under each instance.}
  \label{supple_Result_Figure_Landmark_black}
\end{figure*}

Figure~\ref{supple_Result_Figure_Landmark_black} presents the results with blackhead images as input. Similar to blurhead results, the PDMDec model contributes to the best visual quality. It is worth to note that the smaller landmark error (mean $L_2$ distance) do not mean the better visualization quality. The prediction of face pose and position depends on the body/scene context in the blackhead case. The quality of the generation is evaluated according to the facial organ consistency when the pose and position are reasonable. The mean $L_2$ distance to the detected landmarks (used as \textit{ground truth}) is only a reference.

\begin{figure*}[htp]
  \centering
  \includegraphics[width=0.8\linewidth]{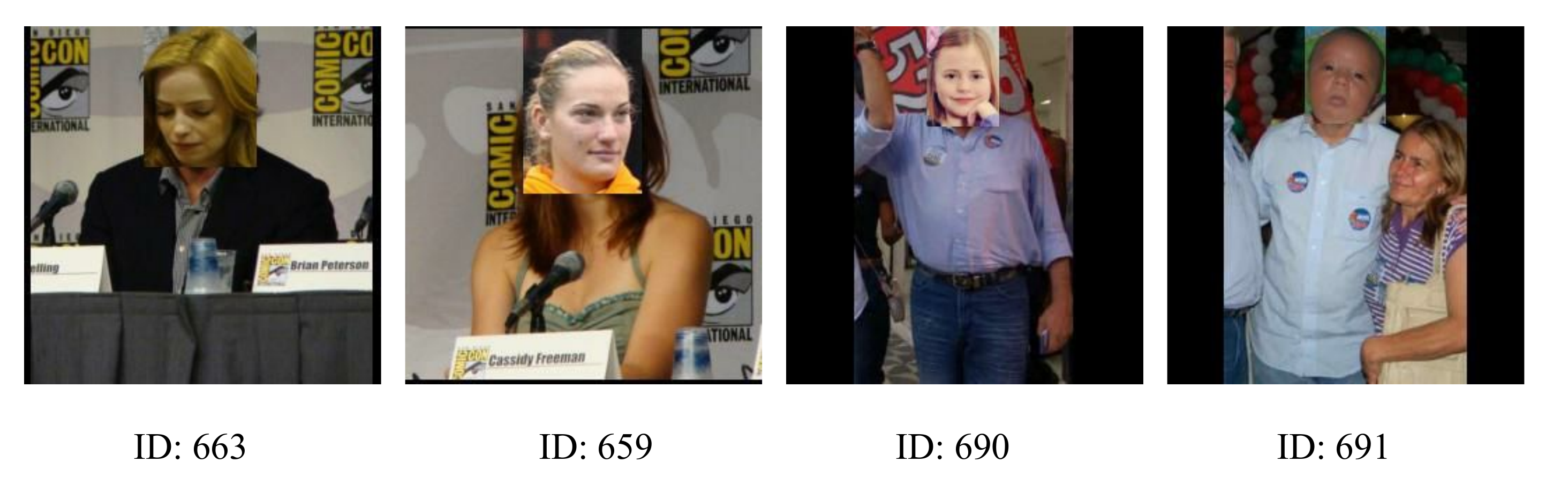}\\
   \vspace{-0.1cm}
     \caption{Visualization results using the direct copy-paste method, corresponding the ``NN head copy-paste'' row in the Table 1 of main paper. Candidate head images are searched in the training data.}
  \label{supple_Result_Figure_copy_paste}
\end{figure*}

Additionally in Figure~\ref{supple_Result_Figure_copy_paste}, we show some examples using direct copy-paste method, corresponding the ``NN head copy-paste'' row in the Table 1 of main paper. Candidate images are searched in the training data based on the normalized $L_2$ distance between detected landmarks. The face poses match the bodies in most cases, but the method results in unpleasant output images. 

\section{Obfuscation performance against AlexNet}
\label{suppsec:alexnet}

Experiments in the main paper have focused on the obfuscation performance with respect to a GoogleNet-based recognizer. However, as argued in the main paper, our obfuscation approach is \emph{target-generic}: it is not generated with respect to a particular recognition system and is expected to work against a generic system.

This section additionally shows the obfuscation performance on an AlexNet-based recognizer. We use the same ``feature extraction - SVM prediction'' framework as in the main paper; we replace the feature extractor by AlexNet. See Table \ref{Supp_Table_Quantitative} for the quantitative comparison between GoogleNet and AlexNet recognizers.

\begin{table*}
\centering
\small
\caption{Evaluation of proposed obfuscation methods against two person recognizers in terms of person recognition rates. This table is an extension of the recognition results in Table 1 of the main paper.
}
\begin{centering}
\begin{tabular*}{17cm}
{ l l l c c c c c c c c }
\multicolumn{3}{c}{Obfuscation method}&&\multicolumn{7}{c}{Obfuscation against person recognizer}\\
\cmidrule{1-3}\cmidrule{5-11}
&\multicolumn{2}{c}{Landmark}&& \multicolumn{3}{c}{GoogleNet} && \multicolumn{3}{c}{AlexNet}\\
\cmidrule{2-3}\cmidrule{5-7}\cmidrule{9-11}
Input &Loss&Decoder  && \texttt{head} & \texttt{body+head} & head contrib. && \texttt{head} & \texttt{body+head} & head contrib.\\
\cmidrule{1-3}\cmidrule{5-7}\cmidrule{9-11}
Original & \multicolumn{2}{l}{No head inpainting}&& 85.6\% & 88.3\% & 72.2\% && 81.6\% & 85.3\% & 66.0\% \\
Original&\multicolumn{2}{l}{NN head copy-paste} && 1.2\% & 7.1\% & 67.5\% && 1.4\% & 6.1\% & 46.2\% \\
\cmidrule{1-3}\cmidrule{5-7}\cmidrule{9-11}
Blur&\multicolumn{2}{l}{No head inpainting} && 52.2\% & 71.6\% & 3.2\% && 52.0\% & 67.0\% & 20.6\% \\
Blur&\multicolumn{2}{l}{Detected landmarks} && 43.7\% & 51.7\% & 70.8\% && 49.0\% & 48.9\% & 37.2\% \\
Blur&$L_2$&Scratch && 36.2\% & 48.4\% & 66.8\% && 44.6\% & 44.6\% & 36.7\% \\
Blur&$L_2$+$D_L$&Scratch && 38.0\% & 48.4\% & 66.6\% && 44.9\% & 45.1\% & 38.9\% \\
Blur&$L_2$+$D_L$&AEDec && 37.5\% & 48.0\% & 66.1\% &&43.9\% &45.0\% &37.5\% \\
Blur&$L_2$+$D_L$&PDMDec && 37.9\% & 49.1\% & 66.7\% && 45.1\% & 45.6\% & 38.0\% \\
\cmidrule{1-3}\cmidrule{5-7}\cmidrule{9-11}
Black&\multicolumn{2}{l}{No head inpainting} && 2.1\% & 67.0\% & 14.0\% && 2.1\% & 63.2\% &1.7\% \\
Black&\multicolumn{2}{l}{Detected landmarks} && 10.1\% & 21.4\%  & 70.8\% && 11.4\% & 20.5\% &46.3\% \\
Black&\multicolumn{2}{l}{NN landmarks} && 7.9\% & 20.4\% & 71.3\% && 10.1\% & 19.0\% & 46.0\% \\
Black&$L_2$&Scratch && 5.8\% & 17.4\% & 73.6\% && 7.5\% & 16.3\% & 49.0\% \\
Black&$L_2$+$D_L$ &Scratch && 5.8\% & 17.2\% & 71.4\% && 7.5\% & 16.4\% &47.4\% \\
Black&$L_2$+$D_L$ &AEDec && 5.6\% & 17.4\% & 72.5\% && 7.5\% & 17.0\% & 48.7\% \\
Black&$L_2$+$D_L$ &PDMDec && 5.6\% & 17.4\% & 71.0\% && 7.4\% & 16.6\% & 51.2\%  \\
\cmidrule{1-3}\cmidrule{5-7}\cmidrule{9-11}
\end{tabular*}
\end{centering}
\vspace{-1mm}
\label{Supp_Table_Quantitative}
\end{table*}

The two recognizers exhibit different behaviours. First of all, on clean images, AlexNet performs worse than GoogleNet ($81.6\%<85.6\%$), while on head-inpainted images, AlexNet shows greater robustness (e.g. $37.9\%$ versus $45.1\%$ on ``Blur input - $L_2+D_L$ - PDMDec''). We also observe systematically less contributions from the head region: $72.2\%$ (GoogleNet) versus $66.0\%$ (AlexNet) on clean images, and consistent drop in head contribution on inpainted images ($20\%\sim 30\%$). AlexNet predictions are supported more by non-head regions, at least partially explaining its robustness against head obfuscation.

Although AlexNet recognizer turns out to behave quite differently from the GoogleNet model, we still reach the same conclusion regarding the superiority of our inpainting-based obfuscation over common patterns like blacking or blurring. For \texttt{body+head}, our inpainting method (``Blur/black input - $L_2+D_L$ - PDMDec'') decreases the recognition rate from $67.0\%$ to $45.6\%$ for blurheads, and from $63.2\%$ to $16.6\%$ for blackheads. Finally, we again observe that the contribution from head region increases as our method inpaints realistic head images. This leads to the same conclusion as for GoogleNet in the main paper: inpainted head images direct recognizer attention to head region, inducing a wrong decision based on the inpainted head.

\end{document}